\documentclass[runningheads]{llncs}
\usepackage[T1]{fontenc}
\usepackage{graphicx}
\usepackage{amsmath}
\usepackage{xcolor}
\usepackage{amssymb}
\usepackage{pifont}
\usepackage{url}
\usepackage[hidelinks]{hyperref}

\begin{document}

\title{Improving the Validity of Automatically Generated Feedback via Reinforcement Learning}

\titlerunning{Improving the Validity of Automatically Generated Feedback}

\author{Alexander Scarlatos\inst{1} \and Digory Smith\inst{2} \and Simon Woodhead\inst{2} \and Andrew Lan\inst{1}}

\authorrunning{A. Scarlatos et al.}

\institute{University of Massachusetts Amherst \and 
Eedi\\
Contact Emails: \email{\{ajscarlatos,andrewlan\}@cs.umass.edu}}

\maketitle

\begin{abstract} 
Automatically generating feedback via large language models (LLMs) in intelligent tutoring systems and online learning platforms has the potential to improve the learning outcomes of many students. However, both feedback generation and evaluation are challenging: feedback content has to be valid especially in subjects like math, which requires models to understand the problem, the solution, and where the student's error lies. Feedback also has to be pedagogically valid to reflect effective tutoring strategies, such as explaining possible misconceptions and encouraging the student, among other desirable features. In this work, we address both problems of automatically generating and evaluating feedback while considering both correctness and alignment. First, we propose a rubric for evaluating math feedback and show that GPT-4 is able to effectively use it to annotate human-written and LLM-generated feedback. Second, we propose a framework for feedback generation that optimizes both correctness and alignment using reinforcement learning (RL). Specifically, we use GPT-4's annotations to create preferences over feedback pairs in an augmented dataset for training via direct preference optimization (DPO). We show that our methods significantly increase the correctness and alignment of generated feedback with Llama 2, an open-source LLM, qualitatively analyze our generation and evaluation systems using case studies, and outline several areas for future work.\footnote{Our code is available at \url{https://github.com/umass-ml4ed/feedback-gen-dpo}.\\Published in AIED 2024: The 25th International
Conference on Artificial
Intelligence in Education.}

\end{abstract}

\section{Introduction}

Providing students with helpful feedback can be critical to their learning, allowing them to quickly address and learn from mistakes. Prior work has shown that delivering immediate automated feedback to students in intelligent tutoring systems and online learning platforms can improve learning outcomes \cite{feedback-effect-2,feedback-effect-1}. However, doing so is challenging since generated feedback should satisfy a wide variety of requirements: it should convey an understanding of the question and why the student's response is incorrect, as well as be aligned with \textit{educational goals} and pedagogical theory. For example, identifying student misconceptions, providing hints, and using encouraging language to promote a growth mindset \cite{growth-mindset,growth-mindset-2} can be helpful, but simply giving away the answer could be detrimental.

Moreover, evaluating generated feedback along these dimensions is also difficult. Automated evaluations must account for both feedback correctness as well as their alignment with educational goals, which requires a thorough understanding of both. Additionally, even when expert-written feedback examples are given as reference, text similarity-based metrics may be unreliable since there are many ways to write valid feedback, and text overlap can emphasize irrelevant features while neglecting more significant ones \cite{socratic,distractors}. While it is common to use human annotators to evaluate feedback, this approach requires significant effort and expenses. Therefore, the lack of reliable, automated feedback \emph{evaluation} methods becomes a bottleneck for developing feedback \emph{generation} methods. 

In this work, we propose a framework that both generates and evaluates feedback messages for incorrect student responses to questions, to improve both their correctness and alignment with educational goals. We ground our work in math education but note that our framework could potentially be generalized to other subjects, such as programming or language learning. First, we propose a \textit{rubric} for evaluating generated feedback and show that LLMs, particularly GPT-4, achieve high agreement with humans in their evaluations.

Second, we use a reinforcement learning (RL)-based approach to generate feedback messages where the reward given to generated feedback during training is based on the evaluation rubric. Moreover, to avoid repeatedly using GPT-4 to evaluate feedback during training, we use direct preference optimization (DPO) \cite{dpo}, an offline RL algorithm, to align the generated feedback with educational goals. This approach is similar to aligning LLMs with human \cite{rlhf} or AI \cite{rlaif} preferences. 
We experiment on a dataset that consists of feedback messages written by math teachers for each incorrect option in multiple-choice questions. Our results show that feedback generated using our framework is significantly more accurate and aligned with educational goals than baselines. Notably, on alignment metrics, we approach the performance of humans and GPT-4, estimated to be a 1T parameter model, using the 7B parameter version of Llama 2.

\section{Related Work}

\subsection{Feedback Generation}

There are many existing approaches for automatic feedback generation. One common method is to use engineered features to detect errors in student responses, and then use a rule-based system to provide relevant feedback or hints \cite{feedback-gen-neil,feedback-effect-2,lan2015mathematical,feedback-effect-1,programming-feedback-ms,programming-feedback-korea}. This method is popular since it is interpretable and reliable but requires significant human effort to adapt to new question types. A recent and more general approach to feedback generation is using large language models (LLMs), either through prompting \cite{socratic,distractors,feedback-gen-decimal,feedback-gen-chatgpt} or fine-tuning \cite{insta-reviewer}. However, prompting pre-trained LLMs requires them to be capable of understanding educational goals, but fine-tuning can yield poor results without significant amounts of aligned training data. We address these concerns in our work by fine-tuning on an augmented dataset annotated with alignment labels.

\subsection{Feedback Evaluation}

Several recent works have used rubrics to evaluate feedback \cite{insta-reviewer,feedback-gen-chatgpt}, and works in other domains have found success in using LLMs to evaluate open-ended text where their judgements correlate with human judgements \cite{llm-eval-open-ended,llm-eval-translation,llm-eval-discourse}. However, most prior works on feedback generation tend to rely on human annotators for reliable evaluation \cite{socratic,eval-bert,insta-reviewer,feedback-gen-chatgpt}. One recent work \cite{feedback-eval-gpt} uses GPT-4 to evaluate math feedback with a rubric and finds high agreement with human annotations. However, they only use GPT-4 to evaluate human-written feedback, while we evaluate feedback written by both humans and LLMs. Including this LLM feedback helps us uncover GPT-4's shortcomings in feedback evaluation, particularly that it can struggle to identify when feedback inaccurately addresses student errors or provides invalid suggestions.

\section{Methodology}

\sloppy
We now detail our framework for the two main tasks of feedback generation and evaluation. Specifically, we first detail our rubric for feedback evaluation and how we collect annotations with GPT-4, followed by how we construct an augmented dataset for training, and finally how we use DPO to fine-tune an LLM for feedback generation. We first define some notations for our tasks. Given a dataset $\mathcal{D}$ of $N$ math questions, we define the $i$-th question as $(q^{(i)}, c^{(i)}, e^{(i)}, \{d^{(i)}_j, f^{(i)}_j | j\in \{1, \ldots, M \} )$. Here, $q^{(i)}$ is the question text, $c^{(i)}$ is the correct answer to the question, $e^{(i)}$ is a textual explanation of the question's solution, $d^{(i)}_j$ is an incorrect, student-generated answer to the question, $f^{(i)}_j$ is a textual feedback message to give to a student when their answer is $d^{(i)}_j$, and $M$ is the number of different incorrect answers given for each question. When discussing individual data points, we omit $i$ and $j$ for notation simplicity. We assume that the feedback messages in the dataset are human-written, and refer to these as the gold, ground-truth feedback. 

\subsection{Feedback Evaluation}

We now detail our rubric for evaluating feedback given to students for their incorrect answers. In addition to correctness, we aim to evaluate feedback messages on their alignment with educational goals, including those associated with a growth mindset \cite{growth-mindset,growth-mindset-2}. We take inspiration from prior works using rubrics for feedback evaluation \cite{insta-reviewer,feedback-gen-chatgpt} and include aspects to target common errors that LLMs make when generating feedback. Specifically, our rubric evaluates feedback on five different aspects, each of them resulting in a binary-valued label:

\begin{itemize}
    \item \textbf{Correct (COR.)} The feedback does not make any incorrect statements and is relevant to the current question and student answer.
    \item \textbf{Revealing (REV.)} The feedback does not directly reveal the correct answer to the student.
    \item \textbf{Suggestion (SUG.)} The feedback provides suggestions to the student that, when followed, will guide them towards the correct answer.
    \item \textbf{Diagnostic (DIA.)} The feedback correctly points out the error the student made or the misconception underlying their answer.
    \item \textbf{Positive (POS.)} The feedback is positive and has an encouraging tone.
\end{itemize}

We now define the \textit{rubric function}, $\operatorname{r}$, which assigns labels to any feedback given a corresponding question and incorrect answer, and a final scalar-valued \textit{rubric score}, $s$, which indicates the feedback's overall quality:
\begin{align*}
    & \operatorname{r}(f|q,c,e,d) = (y_{\text{C}},y_{\text{R}},y_{\text{S}},y_{\text{D}},y_{\text{P}}) = \mathbf{y} \in \{0,1\}^5\\
    & s = y_{\text{C}} \cdot \frac{y_{\text{C}} + y_{\text{R}} + y_{\text{S}} + y_{\text{D}} + y_{\text{P}}}{5} \in [0,1]
\end{align*}
where except for correctness, other rubric aspects are equally weighted. The final rubric score is 0 if the feedback message is incorrect; otherwise, the score increases by increments of $0.2$ for every rubric aspect the feedback satisfies.

While the rubric function can be defined by the output of human annotators, the cost of evaluating feedback using humans is very high, especially when we require frequent evaluation such as during RL training. To address this issue, we use GPT-4, known for its ability to generalize to new tasks, to define a version of the rubric function, $\operatorname{r}_\text{GPT-4}$. Using zero-shot chain-of-thought prompting \cite{zero-shot-cot}, we ask GPT-4 yes or no questions related to each of the 5 labels, and use its output to get an estimated label $\mathbf{y}^\prime$ and corresponding score $s^\prime$. During prompt development, we observed that asking GPT-4 questions performed better than assigning labels based on a formal rubric, that binary labels performed better than a Likert scale, and that asking the negation of the first two questions and flipping the labels after improved accuracy. We leave further exploration of the prompt settings, such as the use of in-context examples, for future work. We provide an example prompt, output, and corresponding labels in Table \ref{tab:prompt-annotation}.

\begin{table}[t]
    \centering
    \small
    \caption{Example prompt, output, and labels for GPT-4 feedback evaluation.}
    \begin{tabular}{|p{1.4cm}|p{10.6cm}|}
        \hline
        \textbf{System} & You are a math education expert. \\
        \hline
        \textbf{Prompt} & Your job is to evaluate feedback given to students on math problems.\newline
Here is the question, the correct solution, the incorrect answer the student gave, and the feedback given to the student:\newline
Question: emma works 36 hours a week. write this as a fraction in its simplest form.
Correct Answer: $\frac{3}{14}$
Solution: There are 24 × 7 = 168 hours in a week, so the fraction is 36/168. This simplifies to 18/84 = 9/42 = 3/14.\newline
Incorrect Answer: $\frac{18}{84}$\newline
Feedback: It looks like you have simplified, but not fully.\newline
For the following questions, provide a short explanation and then answer with "Yes" or "No":\newline
1. Does the feedback make any incorrect statements?\newline
2. Does the feedback directly reveal the answer to the question?\newline
3. Does the feedback give suggestions to the student on how to improve the answer?\newline
4. Does the feedback correctly point out the misconception underlying the student's answer?\newline
5. Does the feedback have a positive or encouraging tone? \\
        \hline
        \textbf{Output} & 1. Explanation: The feedback is correct in stating that the student has simplified, but not fully. The student stopped at $\frac{18}{84}$, which can be further simplified to $\frac{3}{14}$. So, the feedback does not make any incorrect statements. Answer: No.

2. Explanation: The feedback does not provide the correct answer. It only indicates that the student's answer is not fully simplified. Answer: No.

3. Explanation: The feedback does not provide specific suggestions on how to improve the answer. It only states that the student has not fully simplified the fraction. Answer: No.

4. Explanation: The feedback correctly identifies that the student's misconception is not fully simplifying the fraction. Answer: Yes.

5. Explanation: The feedback is neutral in tone. It points out the error but does not use any positive or encouraging language. Answer: No. \\
        \hline
    \textbf{Result} & ($y^\prime_\text{C}=1$, $y^\prime_\text{R}=1$, $y^\prime_\text{S}=0$, $y^\prime_\text{D}=1$, $y^\prime_\text{P}=0$) $\rightarrow$ $s^\prime=0.6$ \\
    \hline
    \end{tabular}
    \label{tab:prompt-annotation}
\end{table}

\subsection{Data Augmentation}

We now detail our method for constructing an augmented dataset, which will be used for RL training as well as calculating agreement between GPT-4 and human annotations. For both of these tasks, we require both \textit{positive} examples, i.e., feedback messages that score highly on the rubric, and \textit{negative} examples, i.e., feedback that score poorly on the rubric. We observe that while most of the gold feedback messages in $\mathcal{D}$ score highly, feedback generated by prompting LLMs can be flawed or unaligned with the evaluation metric, thus scoring poorly. Therefore, we collect 3 LLM-augmented versions of $\mathcal{D}$, where each feedback $f^{(i)}_j$ is replaced with a generated version: $\mathcal{D}_R$, where feedback is generated using few-shot prompting with random in-context examples, $\mathcal{D}_S$, where feedback is generated using few-shot prompting with the most similar examples, and $\mathcal{D}_Z$, where feedback is generated using zero-shot prompting. We refer to the union of the original dataset and LLM-augmented data $\mathcal{D}^\prime=\bigcup\{\mathcal{D},\mathcal{D}_R,\mathcal{D}_S,\mathcal{D}_Z\}$ as the augmented dataset. We use GPT-4 to annotate the feedback messages in this set, and we detail how we use these annotations for training in the next section.

\subsection{Direct Preference Optimization}

In order to generate feedback that scores highly on the rubric, we leverage direct preference optimization (DPO) \cite{dpo}, an offline RL algorithm, due to its simplicity and efficiency. We note that online RL algorithms such as PPO could also apply to our framework, although they would require training a reward model and introduce additional technical challenges due to training instability issues; we leave exploration of such algorithms for future work.
At a high level, DPO trains an LLM on pairs of generated outputs given the same input, where one is preferred over the other. The goal is to use this preference information to make the LLM generate outputs that more closely resemble the preferred outputs seen during training. In our context, the output is the feedback message, $f$, while the input includes the question and incorrect answer information, $x=(q,c,e,d)$.
During training, we minimize the DPO objective, i.e.,
\begin{align*}
    \min_\theta -\mathbb{E}_{(x, f_w, f_l) \sim \mathcal{D}_\text{DPO}}\left[ \log \sigma \left( \beta \log \frac{\pi_\theta(f_w|x)}{\pi_\text{ref}(f_w|x)} - \beta \log \frac{\pi_\theta(f_l|x)}{\pi_\text{ref}(f_l|x)} \right) \right],
\end{align*}
where $f_w$ is preferred over $f_l$ as the feedback for $x$, $\mathcal{D}_\text{DPO}$ is a curated dataset containing these feedback pairs and preferences, $\pi_\theta$ is the trained LLM, i.e., a text generation ``policy'', parameterized by $\theta$, $\pi_\text{ref}$ is a frozen \textit{reference} LLM, and $\beta$ is a hyperparameter to control how far $\pi_\theta$ can deviate from $\pi_\text{ref}$. We now detail how we construct $\mathcal{D}_\text{DPO}$ using both feedback from the augmented dataset and mismatched feedback from the gold dataset.

We first leverage the augmented dataset to construct feedback preference pairs. For each unique $x \in \mathcal{D}^\prime$, we have 4 feedback messages, 1 human and 3 LLM-generated, from which we construct ${4 \choose 2}=6$ unique pairs. We then use the score $s^\prime$ for each feedback to determine which feedback is preferred, and exclude pairs that have the same score. For instance, consider a case where some $x$ has possible feedback messages $f_1$, $f_2$, $f_3$, and $f_4$, with scores $1.0$, $0.8$, $0.4$, and $0.4$, respectively. We then produce the preference pairs $(f_1, f_2)$, $(f_1, f_3)$, $(f_1, f_4)$, $(f_2, f_3)$, and $(f_2, f_4)$, where the first feedback is the preferred one in each pair.

We also use mismatched feedback from the gold dataset to construct additional preference pairs. We observe that feedback written for different incorrect answers to the same question will have many semantically similar features and often the same variables and numbers. However, despite their similarities, the feedback written for the corresponding incorrect answer is almost always better suited than feedback written for other incorrect answers. Therefore, these mismatched feedback are excellent \textit{hard} negatives since it is hard for algorithms to distinguish between them and good feedback; finding such hard negatives has been shown to be the key to contrastive learning \cite{hard-negatives}. In addition to using mismatched feedback from the same question, we construct one more pair using a feedback from a random question in the gold dataset. For instance, for $x^{(i)}_1$ and $M=3$, we construct the preference pairs $(f^{(i)}_1, f^{(i)}_2)$, $(f^{(i)}_1, f^{(i)}_3)$, and $(f^{(i)}_1, f^{(i^\prime)}_{j^\prime})$, for some random $i^\prime \in [1,N]$ and $j^\prime \in [1,M]$, where $f^{(i)}_1$ is preferred in all pairs. 

\section{Experiments}

We now detail all experiments we conduct to validate our framework for feedback generation and evaluation. First, we demonstrate that our methods improve the correctness and alignment of generated feedback using both quantitative evaluation from GPT-4 and qualitative case studies. Second, we demonstrate that GPT-4 has high agreement with human annotations on our rubric, justifying its use as an evaluator, and further investigate its shortcomings using case studies.

\subsection{Dataset}

We validate our framework using a dataset of middle school-level math multiple choice questions from Eedi, a math learning platform. The questions cover a variety of number sense concepts including fractions, exponents, rounding, and many others. All questions and feedback messages are written by real math teachers, deployed to real students, and are generally high quality. There are a total of 1,956 questions in the dataset and each question has a total of 3 incorrect options and a ground truth human-written feedback for each. We remove questions that require images and ones with processing errors, resulting in 1,418 questions. We divide these into a train/validation/test split of 850/284/284 questions and correspondingly 2,550/852/852 incorrect answers and corresponding feedback.

\subsection{Experimental Setting}

\paragraph{Data Augmentation} We use two LLMs to generate feedback for our augmented dataset: \texttt{code-davinci-002} (Codex) \cite{codex} for $\mathcal{D}_R$ and $\mathcal{D}_S$ since it has strong few-shot prompting ability, and \texttt{gpt-3.5-turbo} for $\mathcal{D}_Z$ since its zero-shot ability is much better than \texttt{code-davinci-002}. We use 2 in-context examples for few-shot prompts, only select examples from the train set, and use the S-BERT model \texttt{all-distilroberta-v1} \cite{sbert} to measure similarity for $\mathcal{D}_S$. We prompt the models with questions, correct answers and incorrect answers, but not full solutions to make the task harder and increase the amount of incorrect feedback. To reduce costs, we randomly select a subset of $\mathcal{D}^\prime$ to be annotated by GPT-4. Specifically, we take 10,000, 1,000 and 1,000 samples from the train, validation and test sets, respectively, and remove the remaining samples from the augmented dataset.

\paragraph{Feedback Generation Models}
We primarily use the instruction-tuned Llama-2 7B Chat model \cite{llama-2} from HuggingFace \cite{huggingface} for feedback generation, loaded with 8-bit quantization \cite{quantization}. For both supervised fine-tuning (SFT) and DPO, we train LoRA adapters \cite{lora} on all weight matrices, setting $r=32$, $\alpha=16$, and dropout$=0.05$. We train using the AdamW optimizer with a learning rate of 3e-4 with warmup for 10\% of steps and an effective batch size of 64 using gradient accumulation. We train for 3 epochs, which we find minimizes the loss on the validation set.
For DPO, we set $\beta=0.5$ and use the SFT model for initialization and as the reference model, which empirically outperformed using the base model.
At inference time, we use greedy decoding and set the maximum new tokens to 200.

\paragraph{Metrics} When evaluating feedback, we report the average of each rubric label in $\mathbf{y}^\prime$ and the corresponding scores $s^\prime$ assigned by GPT-4. We note that GPT-4 will very rarely fail to assign labels when feedback is unrelated to the current question, in which case we automatically assign label values of 0. We use a temperature of 0 and 300 maximum tokens for GPT-4 decoding. We also use two popular reference-based metrics with the human-written feedback as reference: ROUGE-L (\textbf{ROU.}) \cite{lin-2004-rouge} which is based on textual overlap, and the F1 of the BERTScore (\textbf{BER.}) \cite{bert-score} using the recommended \texttt{microsoft/deberta-xlarge-mnli} model, which is based on token-level semantic similarity.

\subsection{Feedback Generation}

We now show that we can improve both the correctness and alignment of generated feedback using our framework. We primarily focus on using Llama 2 Chat, an open-source LLM with 7B parameters, where we compare several versions of the model: \textbf{Zero-Shot}, i.e., simply prompting the base LLM, \textbf{SFT}, i.e., fine-tuning the base LLM on the gold feedback set, \textbf{DPO (Score)}, i.e., training the LLM with DPO only on the augmented dataset, and \textbf{DPO (Score + Mismatch)}, i.e., training the LLM with DPO on the augmented dataset and mismatched feedback. We additionally compare with the gold, human-written feedback in the dataset, as well as feedback generated by GPT-4. We use the same prompt for all methods, where we instruct the model to generate short and helpful feedback and to follow a version of the evaluation rubric.

\subsubsection{Quantitative Analysis}

\begin{table*}[t]
    \centering
    \small
    \caption{Quantitative results of feedback generation across methods. Our best method outperforms all Llama 2 baselines in both correctness and alignment.}
    \begin{tabular}{|l|c|c|c|c|c|c|c|c|}
        \hline
        & \textbf{COR.} & \textbf{REV.} & \textbf{SUG.} & \textbf{DIA.} & \textbf{POS.} & \textbf{Score} & \textbf{ROU.} & \textbf{BER.}\\
        \hline
        Human & 0.91 & 0.98 & 0.67 & 0.82 & 0.41 & 0.73 & 1.00 & 1.00\\
        GPT-4 & 0.95 & 0.96 & 0.99 & 0.93 & 1.00 & 0.94 & 0.19 & 0.57\\
        \hline
        Zero-shot & 0.63 & 0.63 & 0.74 & 0.43 & \textbf{1.00} & 0.49 & 0.16 & 0.55 \\
        SFT & 0.65 & \textbf{0.98} & 0.49 & 0.68 & 0.19 & 0.49 & \textbf{0.29} & \textbf{0.61} \\
        DPO (Score) & 0.70 & 0.93 & \textbf{0.95} & 0.82 & 0.66 & 0.65 & 0.22 & 0.57 \\
        DPO (Score + Mismatch) & \textbf{0.77} & 0.96 & \textbf{0.95} & \textbf{0.86} & 0.57 & \textbf{0.71} & 0.23 & 0.57 \\
        \hline
    \end{tabular}
    \label{tab:main-results}
\end{table*}

Table \ref{tab:main-results} shows the average rubric labels and scores assigned by GPT-4 on all feedback in the test set, as well as the ROUGE-L and BERTScore values for reference. We see that DPO (Score + Mismatch) significantly improves the feedback scores compared to baselines (a 45\% increase compared to Zero-Shot and SFT), showing that our data augmentation and training setup is highly effective at improving the quality of feedback generation with Llama 2. We additionally observe that including the mismatched feedback messages substantially increases the correctness of generated feedback, confirming their effectiveness as hard negative examples. Surprisingly, SFT does not outperform Zero-Shot on score, which shows that the standard fine-tuning setup is not effective for feedback generation. We can also see that ROUGE-L and BERTScore are unreliable estimates of feedback quality since they are highest on SFT, primarily because it copies the style of the gold feedback the closest. 

We also see that GPT-4, a much larger model (rumored to have 1T parameters), performs almost perfectly across all labels; DPO (Score + Mismatch) can only match its performance on the revealing and suggestion metrics. However, we note that these results may be inflated, since we also use GPT-4 in evaluation and it is likely to believe that its own generations conform to the rubric. Moreover, we observe that it prefers to be conservative and provides less specific descriptions of student errors, which leads to high scores under our evaluation metric; see below for a detailed example. Nevertheless, we emphasize that a smaller, open-source model is easier for deployment and much cheaper in real-world educational scenarios than a larger, proprietary model. Additionally, we see that the gold, human-written feedback does not score perfectly on correctness, and has a relatively low overall score due to the suggestion and positive metrics; DPO (Score + Mismatch) achieves a similar overall performance. However, the primary reason for the lower human performance is that teachers did not have our evaluation rubrics in mind when they wrote the feedback. 

\subsubsection{Qualitative Analysis}

We also performed qualitative studies to compare the outputs of the different methods and find cases where they succeed or fail; the main findings are: 1) Zero-Shot produces feedback with the right style but struggles to follow instructions, particularly by not identifying the error or revealing the correct answer, and is prone to hallucinations and numerical errors. 2) SFT produces feedback that is generally short and blunt, and usually attempts to identify the error although is often incorrect. 3) DPO (Score) produces feedback that attempts to identify the error, adding details and questions to provide implicit suggestions and increase positivity. It also produces incorrect outputs although less so than SFT. 4) DPO (Score + Mismatch) is more accurate than DPO (Score) in identifying the error. 5) GPT-4 produces feedback with smooth and coherent language but tends to avoid mistakes by not clearly pointing out the error. When it does detail the error, it is often incorrect, although less often than the other methods. 6) For all methods, most failures occur because the LLM cannot correctly identify the error in the incorrect answer. This failure happens more often when multiple reasoning steps are required to arrive at the student's answer. This observation suggests that although state-of-the-art LLMs claim to excel at mathematical reasoning, they cannot conceive how students may make errors \cite{feedback-eval-gpt,misconceptions,distractors}. One solution could be prompting the LLM to evaluate relevant calculations via code execution before writing the feedback \cite{program-of-thought,code-soliloquies}.

To provide more specific examples of several of our findings, we show a representative question and incorrect answer along with corresponding feedback generated by each method in Table \ref{tab:feedback-gradient}. In this example, we see that the human-written feedback directly points out the error in the incorrect answer that the student only considered the denominators, and also suggests simplifying the fraction on the right. GPT-4 prefers to be vague to avoid mistakes by pointing out that the ratio should be simplified and gives a suggestion based on this, but does not explicitly identify why the student's answer was wrong. Zero-Shot does not point out the student's error and directly answers the question, even though it has been instructed to not do so; it then hallucinates an incorrect and irrelevant multiplication before finishing with a positive message. SFT makes a very generic guess at what the error is and does not give a suggestion for improvement. DPO (Score) guesses a more specific but incorrect error, gives a somewhat misleading suggestion, and uses questions to encourage the student. Finally, DPO (Score + Mismatch) correctly identifies the error and gives a helpful suggestion. 

\begin{table*}[t]
    \centering
    \small
    \caption{Feedback generated by all methods for one question and incorrect answer.}
    \scalebox{.95}{
    \begin{tabular}{|p{2cm}|p{10.5cm}|}
        \hline
        \multicolumn{2}{|c|}{\textbf{Input}}\\
        \hline
        \multicolumn{2}{|p{12cm}|}{
            Question: Erica and Rebecca share some marbles in the ratio $ \frac{1}{4}: \frac{6}{8} $. Erica has $\square$ marbles than Rebecca. Which of the following belongs in the gap? \newline Correct Answer: a third as many \newline Incorrect Answer: twice as many
        }\\
        \hline
        \textbf{Source} & \textbf{Feedback}\\
        \hline
        Human & Not quite, the denominator has doubled but the fractions don't have that relationship. Try simplifying 6/8 and then compare it to 1/4 \color{green}\ding{51} \\
        \hline
        GPT-4 & Good effort, but remember to simplify the ratios first. Erica's share is not twice as much, but a fraction of Rebecca's. Try to compare 1/4 to 3/4 again. Keep going! \ding{51} \\
        \hline
        Zero-shot & Great effort! You are on the right track. Since the ratio is 1/4: 3/4, Erica has 1/4, which is a third of 3/4. To find the number of marbles Erica has, you can multiply 3/4 by 3, which gives you 9. So, Erica has 9 marbles. Keep practicing, and you'll get it right next time! \color{red}\ding{55} \\
        \hline
        SFT & I think you have simplified the second fraction incorrectly. \color{red}\ding{55} \\
        \hline
        DPO (Score) & I think you have simplified 6/8 to 1/2. If we simplify 6/8 to 3/4, how many marbles does Erica have? How many does Rebecca have? What is the difference between them? \color{red}\ding{55} \\
        \hline
        DPO (Score + Mismatch) & I think you have found 8/4 = 2. But we need to simplify 6/8 to get the same units as 1/4. Then we can see how many more or less Erica has. \color{green}\ding{51} \\
        \hline
    \end{tabular}}
    \label{tab:feedback-gradient}
\end{table*}

\subsection{Feedback Evaluation}

\begin{table}[t]
    \centering
    \small
    \caption{GPT-4's agreement with human annotations across all rubric labels.}
    \begin{tabular}{|l|c|c|c|c|}
        \hline
        \textbf{Label} & \textbf{Acc.} & \textbf{Prec.} & \textbf{Rec.} & \textbf{F1} \\
        \hline
        COR. & 0.77 & 0.80 & 0.82 & 0.81 \\
        REV. & 0.91 & 0.94 & 0.97 & 0.95 \\
        SUG. & 0.73 & 0.76 & 0.73 & 0.73 \\
        DIA. & 0.68 & 0.56 & 0.85 & 0.68 \\
        POS. & 0.72 & 0.78 & 0.28 & 0.41 \\
        \hline
        Avg. & 0.76 & 0.77 & 0.73 & 0.71 \\
        \hline
    \end{tabular}
    \label{tab:agreement}
\end{table}

Since we use GPT-4 to quantitatively evaluate feedback, we need to verify that GPT-4 can indeed label feedback accurately using our rubric. To do so, we randomly sample 80 feedback messages from the augmented test set and manually evaluate them on the rubric. We also recruited a graduate student with extensive teaching experience to evaluate these feedback messages for an additional set of human annotations. To reduce bias, we do not show human annotators GPT-4's annotations or tell them whether a feedback message is human-written or LLM-generated. We compute agreement between human annotators using the Cohen's kappa statistic, resulting in 0.53, 0.46, 0.35, 0.53 and 0.29 on the COR., REV., SUG., DIA. and POS. labels, respectively. This moderate agreement shows that while humans generally agree, they occasionally make errors due to the difficulty of the task, and have lower agreement on more subjective labels such as POS.

We measure the accuracy (\textbf{Acc.}), precision (\textbf{Prec.}), recall (\textbf{Rec.}), and \textbf{F1} of GPT-4's annotations with respect to human annotations on this set, and report the results averaged across both human annotators in Table \ref{tab:agreement}. We observe that GPT-4 generally agrees with human annotations, with an average accuracy of 76\% across labels. We also compute the Pearson correlation coefficient of the final rubric scores between GPT-4 and human annotations, resulting in 0.56 on average, indicating moderate overall correlation. In contrast, ROUGE-L and BERTScore both have average correlations of 0.40 with human-annotated rubric scores. Not only are these correlations smaller, but they are biased upward since the human-written feedback messages, which generally have high rubric scores, automatically get ROUGE-L and BERTScore values of 1.

\begin{table*}[t]
    \small
    \centering
    \caption{Incorrect GPT-4 rubric annotation examples with explanations.}
    \scalebox{.95}{
    \begin{tabular}{|p{1cm}|p{7cm}|p{4cm}|}
        \hline
        \textbf{Label} & \textbf{Input} & \textbf{Generated Explanation}\\
        \hline
        COR. (FP) & Question: a jacket is reduced in a sale by $20 \%$. a customer has a loyalty card for a further $10 \%$ off the price.  what is the total percentage reduction in the price of the jacket? \newline Correct Answer: $28 \%$ \newline Incorrect Answer: $8 \%$ \newline Feedback: I think you have added the two discounts together. We need to use multiplication to find the overall discount. & The feedback does not make any incorrect statements. It correctly identifies the student's mistake of adding the discounts together instead of multiplying them. \color{red}\ding{55} \\
        \hline
        COR. (FN) & Question: i am thinking of a number. i divide it by 0.5 and my answer is 10. what number am i thinking of? \newline Correct Answer: 5 \newline Incorrect Answer: 20 \newline Feedback: You multiplied by 0.5 instead of dividing by 0.5. & The feedback is incorrect because the student actually multiplied by 2, not 0.5. The correct feedback should be ``You multiplied by 2 instead of dividing by 0.5.'' \color{red}\ding{55} \\
        \hline
        DIA. (FP) & Question: which fraction is equivalent $\frac{3}{7}$ \newline Correct Answer: $\frac{9}{21}$ \newline Incorrect Answer: $\frac{9}{7}$ \newline Feedback: You have multiplied the top and bottom by 3, but you have to multiply by 3 again to get the correct answer & The feedback correctly identifies that the student has not multiplied the denominator by the same number as the numerator, which is the underlying misconception. However, the suggested correction is incorrect. \color{red}\ding{55} \\
        \hline
    \end{tabular}}
    \label{tab:annotation-error-analysis}
\end{table*}

However, GPT-4 still struggles in a few key aspects and we provide examples of erroneous annotations in Table \ref{tab:annotation-error-analysis}. Most importantly, GPT-4 tends to assume that feedback is correct when it sounds convincing but incorrectly identifies the student error or provides an invalid suggestion. These issues mostly occur when calculations are required to verify the feedback. Additionally, GPT-4 can sometimes confuse the roles of variables in the question, leading it to believe that a valid feedback is incorrect. GPT-4 also has a high false positive rate on the diagnostic label due to hallucinating statements that were not made in the feedback. We note that it may be possible to resolve these issues using additional prompt engineering or tools such as self-reflection \cite{reflexion} and code execution to evaluate math expressions \cite{program-of-thought,code-soliloquies}. Finally, while the suggestion and positive labels have relatively low agreement with human annotations, we note that these labels can be very subjective, and that GPT-4's judgement on these labels is more reasonable than these accuracy numbers suggest.

\section{Conclusions and Future Work}

In this work, we proposed a framework for automated feedback generation and evaluation via LLMs for students' incorrect answers in math multiple-choice questions. Our framework accounts for both the mathematical correctness of the feedback and its alignment with good pedagogical practices. We show that using a data augmentation and preference optimization approach, we can generate high-quality feedback using Llama 2 7B, a small and open-source LLM. We also show that GPT-4 can evaluate feedback rather accurately using a rubric and that its annotations are helpful for training the feedback generation method. 
There are many avenues for future work. First, we can apply our framework to other RL algorithms such as PPO, or non-RL approaches such as overgenerate-and-rank. Second, we can evaluate our final feedback generation task via a large-scale human evaluation or classroom study, which would alleviate concerns on GPT-4's annotations being biased. Third, we can test our framework's generalizability by applying it to other domains such as programming or language learning, or other scenarios such as hint generation or student-instructor conversations. Finally, we can consider tailoring feedback to each student according to their knowledge levels \cite{liu2022open}, especially for open-ended questions, since student errors can likely be detected from these responses \cite{mcnichols2023algebra,zhang2022automatic,zhang2021math}. 

\section*{Acknowledgments}
The authors would like to thank Jaewook Lee and Hunter McNichols for assisting in the presentation of this work. The authors also thank Schmidt Futures and the NSF (under grants IIS-2118706 and IIS-2237676) for partially supporting this work.

%
%
%
\bibliographystyle{splncs04}
\bibliography{custom}

\end{document}